\documentclass[runningheads]{llncs}
\usepackage[T1]{fontenc}
\usepackage{graphicx,verbatim}
\usepackage{amsmath}
\usepackage{amssymb}
\usepackage{multirow}
\usepackage{booktabs}
\usepackage{caption}
\usepackage{subcaption}
\usepackage{dirtytalk}
\usepackage{xspace}
\usepackage{arydshln}
\usepackage{bm}
\usepackage{float}
\usepackage{url}
\usepackage[htt]{hyphenat}
\usepackage{kantlipsum}

\DeclareMathOperator{\argmin}{arg\,min}

\newcommand{\seg}{segmentation\xspace}
\newcommand{\segs}{segmentations\xspace}
\newcommand{\anntr}{annotator\xspace}
\newcommand{\anntrs}{annotators\xspace}

\newcommand{\intannt}{inter-annotator\xspace}
\newcommand{\intanntagr}{inter-annotator agreement\xspace}

\newcommand{\isiceight}{ISIC 2018\xspace}
\newcommand{\isicnine}{ISIC 2019\xspace}
\newcommand{\dermpt}{derm7pt\xspace}
\newcommand{\phtwo}{PH\textsuperscript{2}\xspace}
\newcommand{\isicarch}{ISIC Archive\xspace}
\newcommand{\newdatasetnamefull}{ISIC MultiAnnot++\xspace}
\newcommand{\newdatasetname}{IMA++\xspace}
\newcommand{\regmodel}{\mathcal{M}_\textrm{1}}
\newcommand{\diagmodel}{\mathcal{M}_\textrm{2}}
\newcommand{\reghead}{\mathcal{F}_\textrm{R}}
\newcommand{\backbone}{\mathcal{F}_\textrm{F}}
\newcommand{\diaghead}{\mathcal{F}_\textrm{D}}
\newcommand{\regloss}{\mathcal{L}_\textrm{R}}
\newcommand{\diagloss}{\mathcal{L}_\textrm{D}}
\newcommand{\mtmodel}{\mathcal{M}_\textrm{MT}}
\newcommand{\regheadparams}{\mathrm{\Theta}_\textrm{R}}
\newcommand{\backboneparams}{\mathrm{\Theta}_\textrm{F}}
\newcommand{\diagheadparams}{\mathrm{\Theta}_\textrm{D}}

\usepackage[colorlinks]{hyperref}
\newcommand{\ghrepo}{\url{https://github.com/sfu-mial/skin-IAV}}
\newcommand{\rev}[1]{\textcolor{black}{#1}}

\begin{document}
%
% Predicting Inter-Annotator Varability in Skin Lesion Segmentation
% Lesion Segmentation Agreement as a Clinical Feature
% Predicting Lesion Segmentation Agreement Improves Diagnoses
\title{What Can We Learn from Inter-Annotator Variability in Skin Lesion Segmentation?}
\titlerunning{Inter-Annotator Variability in Skin Lesion Segmentation}
%
% \begin{comment}  %% Removed for anonymized MICCAI 2025 submission
\author{
Kumar Abhishek\inst{1}\orcidID{0000-0002-7341-9617} \and
Jeremy Kawahara\inst{2}\orcidID{0000-0002-6406-5300} \and
Ghassan Hamarneh\inst{1}\orcidID{0000-0001-5040-7448}
}
\authorrunning{Abhishek et al.}
\institute{
School of Computing Science, Simon Fraser University, Canada \and
AIP Labs, Budapest, Hungary\\
\email{\{kabhishe,hamarneh\}@sfu.ca, jeremy@aip.ai}
}

% \end{comment}

% \author{Anonymized Authors}  %% Added for anonymized MICCAI 2025 submission
% \authorrunning{Anonymized Author et al.}
% \institute{Anonymized Affiliations \\
%     \email{email@anonymized.com}}

\maketitle              % typeset the header of the contribution
\begin{abstract}
Medical image segmentation exhibits intra- and inter-annot-\\ator variability due to ambiguous object boundaries, annotator preferences, expertise, and tools, among other factors. 
Lesions with ambiguous boundaries, e.g., spiculated or infiltrative nodules, or irregular borders per the ABCD rule, are particularly prone to disagreement and are often associated with malignancy. In this work, we curate \newdatasetname, the largest multi-annotator skin lesion segmentation 
dataset, on which we conduct an in-depth study of variability due to annotator, malignancy, tool, and skill factors. 
We find a statistically significant (\textit{p}$<$0.001) association between \intanntagr (IAA), measured using Dice, and the malignancy of skin lesions. We further show that IAA can be accurately predicted directly from dermoscopic images, achieving a mean absolute error of 0.108. 
Finally, we leverage this association by utilizing IAA as a \say{soft} clinical feature within a multi-task learning objective, yielding a 4.2\% improvement in balanced accuracy averaged across multiple model architectures and across \newdatasetname and four public dermoscopic datasets. The code is available at \ghrepo.

% Authors must provide keywords and are not allowed to remove this Keyword section.

\end{abstract}
\section{Introduction}

Medical image segmentation is a foundational task in modern healthcare, enabling precise quantitative analysis, the development of downstream diagnostic or prognostic models, and treatment planning~\cite{asgari2021deep}. However, the process of delineating structures in medical images, whether performed manually or semi-automatically, is 
prone to variability, leading to intra- and inter-annotator differences. The sources of this variability are multifactorial, including, but not limited to, ambiguous boundaries, varying interpretations of imaging characteristics, discrepancies in annotation protocols, and differences in annotator experience or skill levels. 
In clinical practice, lesions that lack well-defined boundaries and are therefore often difficult to segment, such as spiculated or infiltrative nodules, are often strongly associated with malignancy~\cite{GRIFF2002265,sohns2011value}, suggesting that 
poorly-defined boundaries
may be associated with the underlying disease severity.

Specific to skin image analysis, skin lesion \seg (SLS)\rev{~\cite{celebi2015state,hasan2023survey,mirikharaji2023survey}} can play an important role in computing \seg-based clinical features (e.g.,  irregular borders in the ABCD~\cite{korotkov2012computerized} rule), where the presence of certain clinical features 
can be used to distinguish melanoma from benign lesions and enhance the interpretability of deep learning-based diagnosis methods~\cite{patricio2023coherent,lucieri2022exaid}.
However, reliably computing clinical features derived from lesion \segs can be challenging due to annotator \seg variability. For example, irregular borders or psuedopods, which are clinical features strongly associated with melanoma~\cite{kaya2016abrupt,williams2021assessment}, can be difficult to delineate and may contribute to annotator variability. Thus, in this work, we hypothesize that the level of annotator (dis)agreement in SLS may itself be related to malignancy.
Despite numerous works on modeling annotation styles~\cite{zepf2023label,abhishek2024segmentation,wang2025contour}, \seg selection or aggregation~\cite{ribeiro2020less,mirikharaji2021d,wang2023mse}, and studying variability in expert \segs~\cite{li2010estimating,fortina2012where,ribeiro2019handling} and non-expert annotations of clinical features~\cite{cheplygina2018crowd,raumanns2021enhance}, no prior research has formally investigated if an association exists between the quantitative level of \underline{\textbf{i}}nter-\underline{\textbf{a}}nnotator \seg \underline{\textbf{a}}greement (IAA) and lesion malignancy.

Addressing this gap, we first formally examine whether a systematic relationship exists between IAA levels and lesion malignancy. Using a newly curated dataset, \newdatasetname, we demonstrate a significant
association: malignant lesions exhibit systematically lower levels of IAA compared to benign lesions. 
Based on this observation, we treat the IAA as a type of clinical feature that quantifies how 
ambiguous a lesion is to annotate, which may serve as a proxy for existing clinical features (e.g., irregular border, pseudopods). Driven by this association, our next contribution seeks to predict per-image IAA scores directly from the dermoscopic image using deep regression models, avoiding the segmentation step and allowing us to leverage this signal without requiring multiple annotations during inference.
Finally, motivated by multi-task learning's ability to enhance individual task performance~\cite{maurer2016benefit,zhao2023multi}, and following works that simultaneously predict diagnosis and associated clinical features, like the 7-point criteria to improve diagnostic accuracy and interpretability~\cite{kawahara2018seven,lucieri2022exaid,patricio2023coherent}, our approach views IAA as a \say{soft} clinical feature. Unlike traditional multi-task methods that jointly predict the diagnosis with segmentation~\cite{song2020end,xie2020mutual} or related clinical features such as ABCD~\cite{raumanns2021enhance} (which can be ambiguous due to inter-annotator differences), we hypothesize that training a model to learn the variability in human interpretation 
implicitly captures complex morphological characteristics indicative of malignancy, such as border irregularity and asymmetry, which are often difficult to formalize or are influenced by annotator subjectivity.

\begin{figure*}[ht!]
    \centering
        \includegraphics[width=\textwidth]{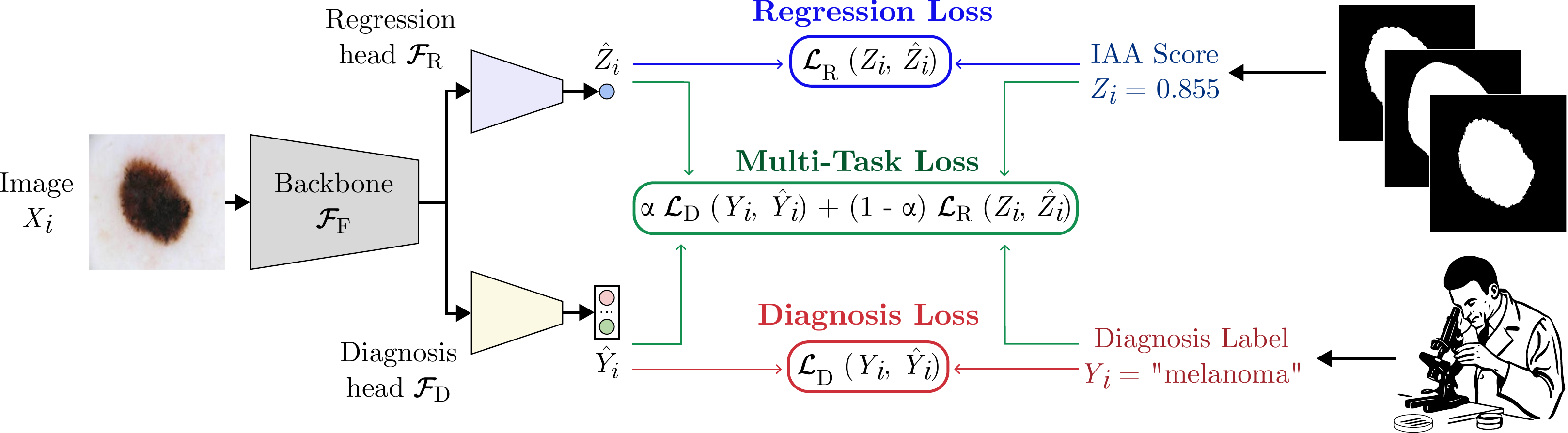}
        \caption{
        Regression ($\regmodel$), diagnosis-only ($\diagmodel$), and multi-task ($\mtmodel$) models.
        }
        \label{fig:overview}
\end{figure*}

To summarize, we make the following contributions: 
\textbf{(1)} We curate, to our knowledge, the largest SLS dataset, \newdatasetname, comprising 5111 masks from 15 unique annotators, and present the largest-scale study of intra- and inter-annota-\\tor variability in this context.
\textbf{(2)} We empirically demonstrate, using rigorous statistical methods, that inter-annotator agreement (IAA) is significantly associated with lesion malignancy in \newdatasetname. 
\textbf{(3)} We show that IAA scores can be predicted, with reasonably low error, directly from image content alone, without requiring any \segs at inference time. 
\textbf{(4)} We demonstrate, through extensive evaluation on multiple datasets, that multi-task models jointly predicting diagnosis and IAA outperform diagnosis-only models.

\section{Methods}
\subsection{Agreement distribution shift across disease classes}

\label{subsec:methods_FOSD_tests}

Let $(\mathcal{X}, \mathcal{Y}, \mathcal{S})$ denote a dataset of $N$ images $\{X_i\}_{i=1}^N$, corresponding $N$ diagnoses $\{Y_i\}_{i=1}^N$, and $N$ sets of multiple segmentation masks $\mathcal{S} = \{\{S_{ik}\}_{k=1}^{K_i}\}_{i=1}^{N}$, where $K_i \geq 2$ is the number of masks for $X_i$. Let $\mathcal{Z} = \{Z_i\}_{i=1}^N$ be the set of corresponding \intanntagr (IAA) scores, where $Z_i = g({\{S_{ik}\}}) \in \mathbb{R}$ is computed per image based on the multiple \segs, where $g(\cdot)$ uses either overlap-based (e.g., Dice similarity coefficient) or boundary-based (e.g., Hausdorff distance) metrics.

First, we wish to rigorously evaluate if there exists a systematic difference between the IAA scores for benign and malignant lesions. 
In particular, we examine the relationship between the probability of sampling a certain value from the IAA distribution of benign versus malignant lesions. To this end, we apply first-order stochastic dominance (FOSD) testing:
a distribution $f_A(x)$ 
is said to first-order stochastically dominate a distribution $f_B(x)$, if $F_A (x) \leq F_B (x) \forall x$, with a strict inequality for some $x$, where $F_A (x)$ and $F_B (x)$ are the cumulative distribution functions (CDFs) of $f_A(x)$ and $f_b (x)$, respectively; 
loosely put, $f_A(x)$ is more likely to generate higher values of $x$ than $f_B(x)$. This first-order stochastic dominance is denoted as $F_A \succeq_1 F_B$.

We define $Z_\textrm{ben}$ and $Z_\textrm{mal}$ as the subsets of IAA scores $\mathcal{Z}$ corresponding to benign and malignant lesions, respectively.
We conduct two separate one-sided tests of FOSD~\cite{barrett2003consistent}: (1) testing whether malignant Dice scores stochastically dominate benign scores, with the hypothesis \( H_{\text{mal} \succeq_1 \text{ben}}: F_\textrm{mal}(x) \leq F_\textrm{ben}(x) \,\, \forall x \), and vice versa  (2) testing whether benign scores stochastically dominate malignant, with the hypothesis  
\( H_{\text{ben} \succeq_1 \text{mal}}: F_\textrm{ben}(x) \leq F_\textrm{mal}(x) \,\, \forall x \).
As a complementary analysis, we also compare the two distributions using a Mann–Whitney \textit{U} test~\cite{mann1947test}.

\subsection{Image-based prediction of \intanntagr}
\label{sec:m1_method}

Next, we examine the ability to predict the IAA score for an image based on the image content alone 
% and without any knowledge of the corresponding \segs.
and without access to the corresponding \segs during inference.
Given an image $X_i$, we predict the target $\hat{Z}_i = \regmodel(X_i; \backboneparams, \regheadparams)$, 
where $\regmodel = \reghead \circ \backbone$, \rev{$\circ$ denotes function composition}, and $\backbone$ and $\reghead$ are the feature-extracting backbone and the regression head, parameterized by $\backboneparams$ and $\regheadparams$, respectively (Fig.~\ref{fig:overview}). The regression model $\regmodel$ is trained by minimizing a regression loss $\regloss$:
\begin{align}
\label{eqn:m1_loss}
    \backboneparams^*, \regheadparams^* = \argmin_{\backboneparams, \regheadparams} \sum_{i=1}^N \regloss(Z_i, \hat{Z}_i).
\end{align}

\subsection{Integrating \intanntagr and diagnosis prediction}
\label{subsec:mtmodel_method}

Given an image $X_i$, a typical image-based diagnosis model $\diagmodel$ predicts \\$\hat{Y}_i = \diagmodel(X_i; \backboneparams, \diagheadparams)$,
where $\diagmodel = \diaghead \circ \backbone$, and $\diaghead$ is the diagnosis head parameterized by $\diagheadparams$. The diagnosis model is optimized by minimizing a diagnosis (classification) loss $\diagloss$:
\begin{equation}
\label{eqn:m2_loss}
    \backboneparams^*, \diagheadparams^* = \argmin_{\backboneparams, \diagheadparams} \sum_{i=1}^N \diagloss(Y_i, \hat{Y}_i).
\end{equation}

Finally, inspired by previous works on multi-task learning in medical imaging~\cite{zhao2023multi} and skin images in particular~\cite{yang2017novel,kawahara2018seven,song2020end,xie2020mutual,abhishek2021predicting,raumanns2021enhance}, we investigate whether simultaneous prediction of IAA and diagnosis improves the accuracy of the latter. To this end, we train $\mtmodel$ to simultaneously predict $\hat{Y}_i$ and $\hat{Z}_i$ such that $(\hat{Y}_i, \hat{Z}_i) = \mtmodel (X_i; \backboneparams, \diagheadparams, \regheadparams)$,
where $\mtmodel = (\reghead \circ \backbone, \diaghead \circ \backbone)$ is a multi-task prediction model with prediction heads for diagnosis (classification) and IAA score (regression) that share the same backbone (Fig.~\ref{fig:overview}). $\mtmodel$ is trained by minimizing a (weighted) sum of the two tasks' objectives:
\begin{equation}
\label{eqn:mtmodel_loss}
    \backboneparams^*, \diagheadparams^*, \regheadparams^* = \argmin_{\backboneparams, \diagheadparams, \regheadparams} \sum_{i=1}^N \left[ \alpha \cdot \diagloss(Y_i, \hat{Y}_i) + (1-\alpha) \cdot \regloss(Z_i, \hat{Z}_i) \right],
\end{equation}
where $\alpha$ is a loss-weighting hyperparameter. Note that $\alpha=0$ and $\alpha=1$ are equivalent to regression-only ($\regmodel$) and diagnosis-only ($\diagmodel$) models, respectively.
More details about exact model architectures, losses \rev{(Eqns.~\ref{eqn:m1_loss}, \ref{eqn:m2_loss}, \ref{eqn:mtmodel_loss})}, datasets, training, and evaluation are discussed in the next section.

\section{Results and Discussion}

\subsection{Datasets and analysis}

\begin{figure}[htbp!]
    \centering
    \includegraphics[width=\linewidth]{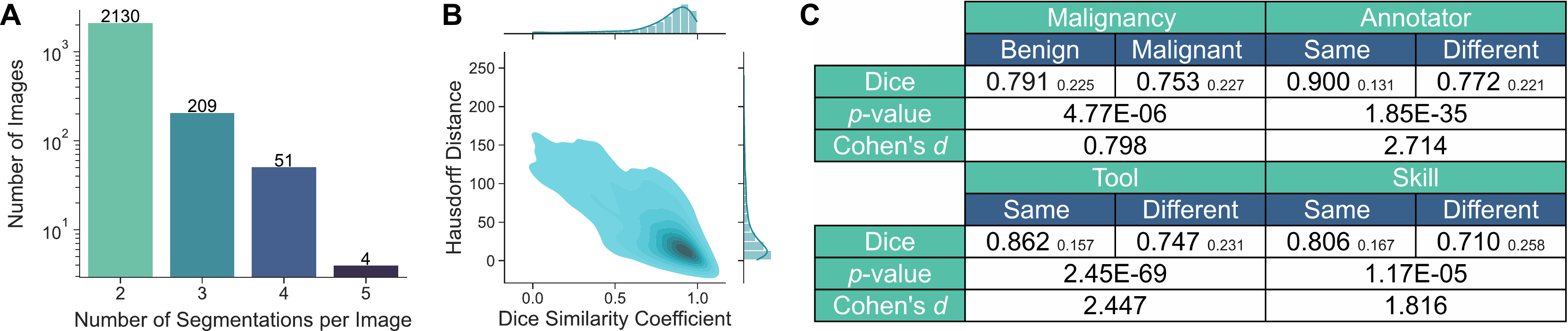}
    \caption{
    \newdatasetname dataset statistics: (A) number of \segs per image, (B) pairwise agreement metrics (Dice and Hausdoff distance), (C) intra- and inter-factor agreement (mean\tiny{std. dev.}\normalsize~ of Dice) with \textit{p}-value and Cohen's \textit{d}. 
    }
    \label{fig:dataset_statistics}
\end{figure}

\textbf{A new curated dataset:} Prior work on multi-\anntr skin lesion segmentation has produced either 
\textbf{(a)} large datasets without annotator-level information (e.g., Ribeiro et al.~\cite{ribeiro2019handling,ribeiro2020less}, Mirikharaji et al.~\cite{mirikharaji2021d}: 2223 images, 4647 total \segs)
or,
\textbf{(b)} small datasets with annotator or style metadata (e.g., Zepf et al.~\cite{zepf2023label}: 100 images, 300 total \segs, 3 styles; Abhishek et al.~\cite{abhishek2024segmentation}: 454 images, 1058 total \segs, 10 unique annotators). 
In this work, we curate and publicly release a new dataset from the \isicarch, called \textbf{\newdatasetnamefull} (\textbf{\newdatasetname} hereafter). 
\rev{
It contains 2394 dermoscopic images segmented by 15 unique annotators, where 2130 images have 2 masks, 209 images have 3 masks, 51 images have 4 masks, and 4 images have 5 masks, resulting in a total of 5111 segmentation masks (Fig.~\ref{fig:dataset_statistics}).
}
% It contains 2394 dermoscopic images segmented by 15 unique annotators, 
% \rev{with 2130, 209, 51, 4 images containing 2, 3, 4, 5 masks each, respectively,}
% resulting in 5111 segmentation masks (\rev{($2130 \times 2) + (209 \times 3) + (51 \times 4) + (4 \times 5) = 5111$}; Fig.~\ref{fig:dataset_statistics}A). 
To the best of our knowledge, \newdatasetname is the largest public multi-\anntr skin lesion segmentation dataset in terms of both mask and annotator counts.

Each mask contains information about the tool used: (T1) manual polygon tracing by a human expert, (T2) semi-automated flood-fill with expert-defined parameters, or (T3) a fully-automated segmentation reviewed and accepted by a human expert; and the skill level of the manual reviewer: (S1) expert or (S2) novice. We partition the images in \newdatasetname into training, validation, and testing splits in the ratio of 70:15:15, stratified by malignancy, number of \segs per image, and Dice score range: low ($<0.5)$, medium, and high ($> 0.8)$.

\smallskip

\noindent\textbf{Calculation of IAA scores:} All images and binary \segs are resized to $256 \times 256$. 
For each image $X_i$, we compute the Dice and Hausdorff distance between all $\binom{K_i}{2}$ unique pairs of segmentation masks (Fig.~\ref{fig:dataset_statistics}B shows the full distribution of pairwise scores).
Although previous IAA studies have used Cohen's kappa~\cite{ribeiro2020less} and Fleiss' kappa~\cite{abhishek2024segmentation}, these metrics measure categorical agreement and fail to capture spatial overlap between annotations; thus, we adopt the Dice metric, which is standard in IAA studies in medical imaging~\cite{sampat2006measuring,ahamed2023comprehensive,gut2024use}.
For each image, we average the pairwise Dice scores to obtain a single IAA score.
Although most lesions tend to exhibit high agreement between \anntrs ([344, 818] out of 2394 images have Dice above [0.95, 0.90]), a notable subset shows poor agreement (23 images have 0 Dice), highlighting the wide range of inter-annotator agreement, and in line with the previous study by Ribeiro et al.~\cite{ribeiro2019handling}.

\begin{figure}[htbp!]
    \centering

    % Start first minipage (for table)
    \begin{subfigure}[b]{0.615\textwidth}
        \centering
        % \begin{table}[H]
% \centering
% \caption{Quantitative results (mean\tiny{\emph{std.dev.}}\normalsize) for predicting the \intanntagr using $\regmodel$. \textsuperscript{\textdagger}\xspace denotes top 3 models by overall MAE.}
% \label{tab:M1_results}
\resizebox{\textwidth}{!}{%
\begin{tabular}{@{}lccccc:ccc:c@{}}
\toprule
\multirow{2}{*}{\textbf{Model}} & \multirow{2}{*}{\textbf{\begin{tabular}[c]{@{}c@{}}Params\\ (M)\end{tabular}}} & \multirow{2}{*}{\textbf{\begin{tabular}[c]{@{}c@{}}MACs\\ (G)\end{tabular}}} & \multicolumn{3}{c}{\textbf{MAE}}                                & \multicolumn{3}{c}{\textbf{MSE}}                                & \multirow{2}{*}{\textbf{\textit{p}-value}} \\ \cmidrule(lr){4-9}
                                &                                                                                    &                                                                                   & Benign              & Malignant           & Overall             & Benign              & Malignant           & Overall             &                                                                                        \\ \midrule % \cmidrule(r){1-3} \cmidrule(lr){6-6} \cmidrule(l){9-10} 

%%%%%%%%%%%%%%%% 3-decimal digits, M params, G MACs, and MW test %%%%%%%%%%%%%%%%
\textbf{VGG-16}                 & 14.72                              & 15.36                       & 0.118\tiny{\emph{0.166}} & 0.134\tiny{\emph{0.188}} & 0.121\tiny{\emph{0.171}} & 0.027\tiny{\emph{0.059}} & 0.035\tiny{\emph{0.077}} & 0.029\tiny{\emph{0.064}} & 1.54E-05       \\
\textbf{ResNet-18}\textsuperscript{\textdagger}        & 11.31       & 1.81                        & 0.103\tiny{\emph{0.158}} & 0.127\tiny{\emph{0.178}} & 0.108\tiny{\emph{0.162}} & 0.025\tiny{\emph{0.065}} & 0.032\tiny{\emph{0.062}} & 0.026\tiny{\emph{0.064}} & 1.15E-08       \\
\textbf{ResNet-50}              & 24.03                              & 4.09                        & 0.124\tiny{\emph{0.175}} & 0.143\tiny{\emph{0.193}} & 0.128\tiny{\emph{0.180}} & 0.031\tiny{\emph{0.074}} & 0.038\tiny{\emph{0.081}} & 0.032\tiny{\emph{0.076}} & 1.41E-37       \\
\textbf{MobileNetV2}\textsuperscript{\textdagger}      & 2.55        & 0.30                        & 0.103\tiny{\emph{0.157}} & 0.129\tiny{\emph{0.182}} & 0.109\tiny{\emph{0.163}} & 0.025\tiny{\emph{0.067}} & 0.033\tiny{\emph{0.070}} & 0.026\tiny{\emph{0.068}} & 3.15E-15       \\
\textbf{MobileNetV3L}           & 3.22                               & 0.21                        & 0.106\tiny{\emph{0.156}} & 0.131\tiny{\emph{0.183}} & 0.111\tiny{\emph{0.162}} & 0.024\tiny{\emph{0.063}} & 0.033\tiny{\emph{0.070}} & 0.026\tiny{\emph{0.065}} & 6.53E-09       \\
\textbf{DenseNet-121}           & 7.22                               & 2.83                        & 0.131\tiny{\emph{0.182}} & 0.141\tiny{\emph{0.191}} & 0.133\tiny{\emph{0.184}} & 0.033\tiny{\emph{0.074}} & 0.037\tiny{\emph{0.082}} & 0.034\tiny{\emph{0.076}} & 5.21E-32       \\
\textbf{EfficientNet-B0}        & 4.34                               & 0.38                        & 0.110\tiny{\emph{0.164}} & 0.138\tiny{\emph{0.191}} & 0.116\tiny{\emph{0.170}} & 0.027\tiny{\emph{0.068}} & 0.036\tiny{\emph{0.076}} & 0.029\tiny{\emph{0.070}} & 1.01E-15       \\
\textbf{EfficientNet-B1}\textsuperscript{\textdagger}  & 6.84        & 0.57                        & 0.107\tiny{\emph{0.165}} & 0.121\tiny{\emph{0.177}} & 0.110\tiny{\emph{0.167}} & 0.027\tiny{\emph{0.074}} & 0.032\tiny{\emph{0.088}} & 0.028\tiny{\emph{0.077}} & 1.63E-07       \\
\textbf{ConvNeXt-T}             & 28.02                              & 4.47                        & 0.130\tiny{\emph{0.195}} & 0.155\tiny{\emph{0.207}} & 0.135\tiny{\emph{0.199}} & 0.039\tiny{\emph{0.103}} & 0.048\tiny{\emph{0.100}} & 0.041\tiny{\emph{0.102}} & 2.62E-28      \\
\textbf{Swin-T}                 & 27.72                              & 4.50                        & 0.131\tiny{\emph{0.188}} & 0.152\tiny{\emph{0.203}} & 0.135\tiny{\emph{0.192}} & 0.035\tiny{\emph{0.089}} & 0.043\tiny{\emph{0.092}} & 0.037\tiny{\emph{0.090}} & 3.70E-37      \\
\textbf{SwinV2-T}               & 27.78                              & 5.96                        & 0.127\tiny{\emph{0.195}} & 0.155\tiny{\emph{0.207}} & 0.133\tiny{\emph{0.198}} & 0.039\tiny{\emph{0.105}} & 0.048\tiny{\emph{0.102}} & 0.041\tiny{\emph{0.104}} & 6.02E-19      \\
\textbf{ViT-B/16}               & 86.00                              & 16.86                       & 0.122\tiny{\emph{0.179}} & 0.149\tiny{\emph{0.206}} & 0.128\tiny{\emph{0.186}} & 0.032\tiny{\emph{0.082}} & 0.044\tiny{\emph{0.102}} & 0.035\tiny{\emph{0.086}} & 3.35E-22       \\
\textbf{ViT-B/32}               & 87.65                              & 4.37                        & 0.129\tiny{\emph{0.181}} & 0.149\tiny{\emph{0.202}} & 0.133\tiny{\emph{0.186}} & 0.033\tiny{\emph{0.080}} & 0.041\tiny{\emph{0.091}} & 0.035\tiny{\emph{0.083}} & 5.94E-44       \\
%%%%%%%%%%%%%%%%%%%%%%%%%%%%%%%%%%%%%%%%%%%%%%%%%%%%%%%%%%%%%%%%

\bottomrule
\end{tabular}%
}
\caption{
% MAE, MSE (mean\tiny{\emph{std.dev.}}\normalsize), and $p$-values for IAA prediction for 13 model architectures ($\regmodel$).\\\textsuperscript{\textdagger}\xspace denotes top 3 models by overall MAE.
}
% \end{table} % includes the external table
        \label{tab:M1_results}
    \end{subfigure}
    \hfill
    % Start second minipage (for figure)
    \begin{subfigure}[b]{0.36\textwidth}
        \centering
        \includegraphics[width=0.9\linewidth]{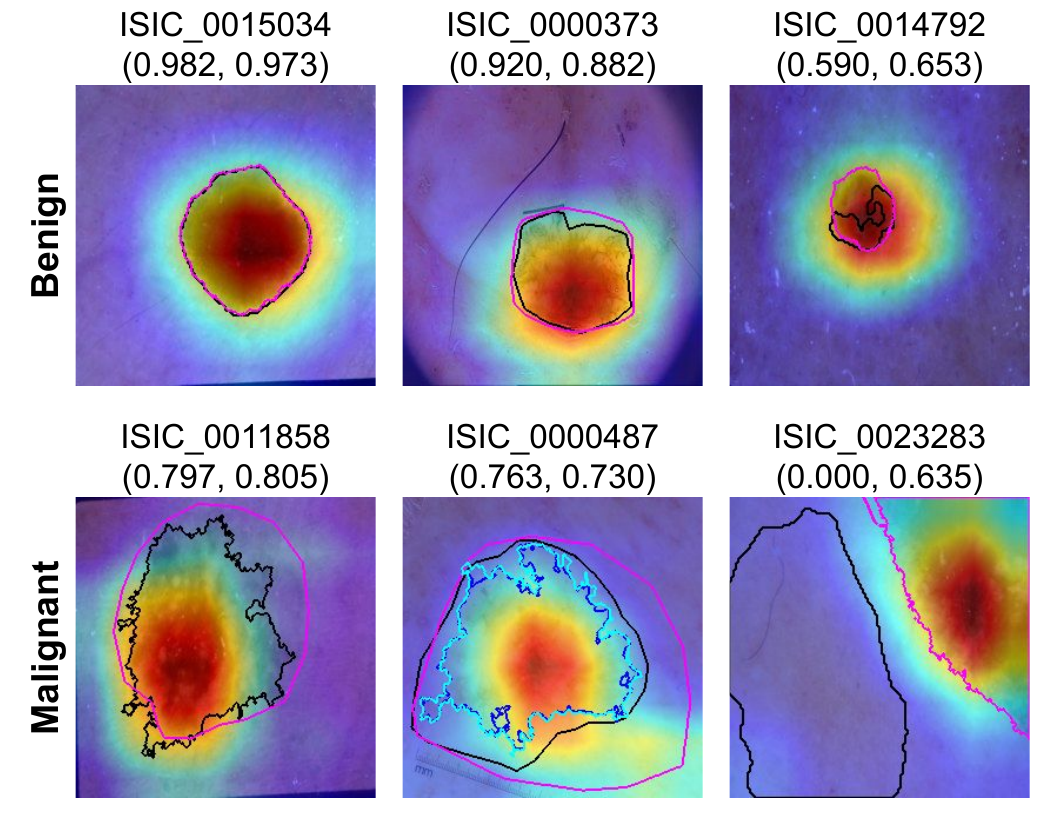}
        \caption{
        % Test images' GradCAM++ outputs from $\regmodel$ (ResNet-18).
        }
        \label{fig:gradcam++}
    \end{subfigure}
    \caption{Predicting \intanntagr (Dice) on the proposed \newdatasetname dataset. (a) Quantitative results (mean\tiny{\emph{std.dev.}}\normalsize of MAE and MSE, and $p$-values for 13 model architectures ($\regmodel$). \textsuperscript{\textdagger}\xspace denotes top 3 models by overall MAE. (b) GradCAM++ saliency heatmaps from $\regmodel$ (ResNet-18). 
    Each image $X_i$ shows
    corresponding overlaid \segs $\{S_{ik}\}_{k=1}^{K_i}$ and
    (ground truth IAA $Z_i$, predicted IAA $\hat{Z}_i$) below the ISIC image ID.
    }

\end{figure}
% MAE, MSE (mean\tiny{\emph{std.dev.}}\normalsize), and $p$-values for IAA prediction for 13 model architectures ($\regmodel$).\\\textsuperscript{\textdagger}\xspace denotes top 3 models by overall MAE.

\smallskip

\noindent\textbf{Is malignancy associated with IAA?} Fig.~\ref{fig:dataset_statistics}C reveals a notable difference in IAA scores between benign and malignant lesions: benign lesions tend to exhibit higher Dice scores (0.791 $\pm$ 0.215 vs. 0.753 $\pm$ 0.227). However, a comparison of means alone can be misleading if the underlying distributions differ in shape or variance. To address this, we compared the full distributions of IAA scores for benign and malignant lesions using the Mann-Whitney \textit{U} test, which confirmed that agreement is significantly higher for benign lesions ($p < 0.01$), suggesting greater \anntr consensus in those cases.

Our FOSD tests (Sec.~\ref{subsec:methods_FOSD_tests}), conducted using PySDTest~\cite{lee2023pysdtest}, reinforce this conclusion. 
Using 1,000 bootstrap resampling iterations at significance level $\alpha = 0.001$, we rejected the hypothesis that malignant lesions stochastically dominate benign ones ($H_{\text{mal} \succeq_1 \text{ben}}$, $p < 0.001$), while the reverse hypothesis ($H_{\text{ben} \succeq_1 \text{mal}}$) was not rejected ($p = 0.923$).
Together, these results 
% provide strong statistical evidence 
support 
that the distribution of \intannt agreement for benign lesions first-order stochastically dominates that for malignant lesions, indicating 
higher segmentation consensus for benign cases. 
This 
is likely due to benign lesions often exhibiting
more well-defined, homogeneous boundaries, making them easier to segment consistently. In contrast, malignant lesions tend to be more heterogeneous in appearance and morphology, which likely contributes to
higher annotation variability.

\smallskip

\noindent\textbf{Impact of other annotation factors on IAA:} In addition to malignancy, Fig.~\ref{fig:dataset_statistics}C summarizes intra- and inter-factor-dependent IAA scores, along with corresponding Mann–Whitney \textit{U} test \textit{p}-values and Cohen's \textit{d}~\cite{cohen2013statistical} effect sizes. As expected, and consistent with findings in other medical imaging modalities~\cite{fu2014interrater,gut2024use}, intra-annotator agreement is significantly higher than inter-annotator agreement. We also observe that segmentations performed using the same annotation tool tend to show higher agreement. Similarly, annotators with the same skill level exhibit greater consistency, particularly in the case of malignant lesions.
To our knowledge, this represents the largest study of annotator variability in skin lesion segmentation to date in terms of dataset size, substantially exceeding the scale of prior work~\cite{li2010estimating,fortina2012where,peruch2013simpler,tschandl2019domain,ribeiro2019handling,ribeiro2020less}.

\begin{table}[ht!]
\centering
\caption{Comparing the diagnostic performance of $\diagmodel$ to $\mtmodel$ on \newdatasetname for different values of $\alpha$ (Eqn.~\ref{eqn:mtmodel_loss}). $\alpha=0.9$ performs the best across all architectures.}
\label{tab:ablation_results}
\resizebox{0.9\textwidth}{!}{%
\setlength{\tabcolsep}{0.75em}
\def\arraystretch{1.35}
\begin{tabular}{@{}cccc:cc:cc@{}}
\toprule
\multicolumn{2}{c}{\multirow{2}{*}{}}                                                                       & \multicolumn{2}{c}{ResNet-18}             & \multicolumn{2}{c}{MobileNetV2}           & \multicolumn{2}{c}{EfficientNet-B1}       \\ % \cmidrule(l){3-8} 
\multicolumn{2}{c}{}                                                                                        & Bal. Acc.           & AUROC               & Bal. Acc.           & AUROC               & Bal. Acc.           & AUROC               \\ \midrule % \cmidrule(r){1-2}

%%%% 3 digits with tiny-emph

\multicolumn{2}{c}{\begin{tabular}[c]{@{}c@{}}Diagnosis Only\\ ($\diagmodel$)\end{tabular}}                         & 0.746\tiny{\emph{0.008}} & 0.835\tiny{\emph{0.003}} & 0.757\tiny{\emph{0.009}} & 0.843\tiny{\emph{0.004}} & 0.746\tiny{\emph{0.009}} & 0.827\tiny{\emph{0.001}} \\ \hdashline
\multirow{5}{*}{\begin{tabular}[c]{@{}c@{}}Multi-\\ Task\\ Learning\\ ($\mtmodel$)\end{tabular}} & $\alpha = 0.1$ & 0.711\tiny{\emph{0.009}} & 0.785\tiny{\emph{0.001}} & 0.748\tiny{\emph{0.003}} & 0.859\tiny{\emph{0.001}} & 0.744\tiny{\emph{0.016}} & 0.826\tiny{\emph{0.018}} \\
                                                                                           & $\alpha = 0.2$ & 0.723\tiny{\emph{0.009}} & 0.822\tiny{\emph{0.002}} & 0.740\tiny{\emph{0.007}} & 0.857\tiny{\emph{0.035 }} & 0.750\tiny{\emph{0.000}} & 0.853\tiny{\emph{0.002}} \\
                                                                                           & $\alpha = 0.5$ & 0.750\tiny{\emph{0.004}} & 0.852\tiny{\emph{0.006}} & 0.785\tiny{\emph{0.006}} & 0.869\tiny{\emph{0.006}} & 0.738\tiny{\emph{0.010}} & 0.869\tiny{\emph{0.003}} \\
                                                                                           & $\alpha = 0.8$ & 0.757\tiny{\emph{0.004}} & 0.852\tiny{\emph{0.001}} & 0.797\tiny{\emph{0.011}} & 0.879\tiny{\emph{0.002}} & 0.767\tiny{\emph{0.007}} & 0.873\tiny{\emph{0.001}} \\
                                                                                           & $\alpha = 0.9$ & 0.765\tiny{\emph{0.002}} & 0.869\tiny{\emph{0.002}} & 0.805\tiny{\emph{0.004}} & 0.882\tiny{\emph{0.001}} & 0.772\tiny{\emph{0.009}} & 0.878\tiny{\emph{0.003}} \\
                                                                                           
\bottomrule
\end{tabular}%
}
\end{table}

\smallskip

\noindent\textbf{Other datasets:} In addition to \newdatasetname, we also conduct experiments on 4 other dermoscopic image datasets: \phtwo~\cite{ballerini2013color}, \dermpt~\cite{kawahara2018seven}, \isiceight~\cite{codella2019skin,tschandl2018ham10000}, \isicnine~\cite{codella2018skin,tschandl2018ham10000,hernandez2024bcn20000}. We use the standardized
partitions for ISIC 2018, 2019, and split \phtwo and \dermpt into train:valid:test in
70:15:15 ratio stratified by diagnosis.

All models were trained on an Ubuntu 20.04 workstation with AMD Ryzen 9 5950X, 32 GB RAM, NVIDIA RTX 3090 with Python 3.10.18 and PyTorch 2.7.1. All reported metrics are mean\tiny{\emph{std. dev.}} \normalsize over 3 runs with different seeds. All trained models and code 
\rev{are available}
at \ghrepo.

\subsection{Image-based prediction of \intanntagr}
\label{subsec:m1_results}

To directly predict IAA scores from images (Sec.~\ref{sec:m1_method}), we evaluate 13 architectures spanning CNNs and Transformers, 
covering
a wide range of capacities in terms of parameters and  multiply-accumulate operations (MACs).
Each model uses the backbone as a feature extractor with a regression head: \texttt{Linear(256)} $\to$ \texttt{BatchNorm1D} $\to$ \texttt{ReLU} $\to$ \texttt{Dropout(0.5)} $\to$ \texttt{Linear(1)}. All models were trained for 50 epochs using SGD (momentum = 0.9, weight decay = 1e-4, batch size = 32, learning rate = 1e-2 decayed $\times 0.1$ every 10 epochs). We use Smooth-$L_1$ loss~\cite{girshick2015fast} as $\regloss$, selecting the model with the lowest validation MAE. Results are reported in terms of MAE, MSE, and Mann–Whitney \textit{U} test \textit{p}-values (Fig.~\ref{tab:M1_results}).

All models achieve good predictive performance (MAE $\in [0.10, 0.135]$), suggesting that IAA scores can be inferred from image content alone. 
Grad-CAM++ \cite{chattopadhay2018grad} visualizations (Fig.~\ref{fig:gradcam++}) for the best model (ResNet-18) confirm saliency focused on the lesions and their boundaries. Notably, the third malignant example shows the model correctly localizing the lesion and predicting a plausible IAA (0.635), despite the \say{true} IAA being 0.0, highlighting label noise rather than prediction error.
For all subsequent analyses, we use the top 3 performing architectures: ResNet-18, MobileNetV2, and EfficientNet-B1.

\begin{table}[ht!]
\centering
\caption{Evaluating generalization performance on four other dermoscopic image datasets with 3 model architectures ($\mtmodel$ with $\alpha$ set to 0.9 based on Table~\ref{tab:ablation_results}).}
\label{tab:other_datasets}
\resizebox{0.9\textwidth}{!}{%
\setlength{\tabcolsep}{0.7em}
\def\arraystretch{1.35}
\begin{tabular}{@{}cccc:cc:cc@{}}
\toprule
\multicolumn{2}{c}{\multirow{2}{*}{}}                                                     & \multicolumn{2}{c}{ResNet-18}             & \multicolumn{2}{c}{MobileNetV2}           & \multicolumn{2}{c}{EfficientNet-B1}       \\ % \cmidrule(l){3-8} 
\multicolumn{2}{c}{}                                                                      & Bal. Acc.           & AUROC               & Bal. Acc.           & AUROC               & Bal. Acc.           & AUROC               \\ \midrule % \cmidrule(r){1-2}

%%%% 3 digits with tiny-emph

\multirow{2}{*}{\phtwo}                                                  & Diag. Only ($\diagmodel$) & 0.938\tiny{\emph{0.000}} & 0.988\tiny{\emph{0.000}} & 0.943\tiny{\emph{0.033}} & 0.988\tiny{\emph{0.007}} & 0.870\tiny{\emph{0.009}} & 0.979\tiny{\emph{0.002}} \\
                                                                      & Multi-Task ($\mtmodel$) & 0.979\tiny{\emph{0.009}} & 0.992\tiny{\emph{0.000}} & 0.979\tiny{\emph{0.009}} & 0.999\tiny{\emph{0.002}} & 0.964\tiny{\emph{0.009}} & 0.984\tiny{\emph{0.004}} \\ \hdashline
\multirow{2}{*}{\dermpt}                                              & Diag. Only ($\diagmodel$) & 0.734\tiny{\emph{0.009}} & 0.836\tiny{\emph{0.009}} & 0.654\tiny{\emph{0.007}} & 0.800\tiny{\emph{0.003}} & 0.756\tiny{\emph{0.037}} & 0.862\tiny{\emph{0.015}} \\
                                                                      & Multi-Task ($\mtmodel$) & 0.748\tiny{\emph{0.005}} & 0.846\tiny{\emph{0.001}} & 0.792\tiny{\emph{0.012}} & 0.887\tiny{\emph{0.002}} & 0.774\tiny{\emph{0.011}} & 0.861\tiny{\emph{0.003}} \\ \hdashline
\multirow{2}{*}{\begin{tabular}[c]{@{}c@{}}ISIC \\ 2018\end{tabular}} & Diag. Only ($\diagmodel$) & 0.744\tiny{\emph{0.005}} & 0.893\tiny{\emph{0.002}} & 0.727\tiny{\emph{0.007}} & 0.872\tiny{\emph{0.000}} & 0.713\tiny{\emph{0.066 }} & 0.868\tiny{\emph{0.002}} \\
                                                                      & Multi-Task ($\mtmodel$) & 0.752\tiny{\emph{0.003}} & 0.898\tiny{\emph{0.001}} & 0.745\tiny{\emph{0.007}} & 0.903\tiny{\emph{0.003}} & 0.753\tiny{\emph{0.012}} & 0.885\tiny{\emph{0.047}} \\ \hdashline
\multirow{2}{*}{\begin{tabular}[c]{@{}c@{}}ISIC\\ 2019\end{tabular}}  & Diag. Only ($\diagmodel$) & 0.670\tiny{\emph{0.004}} & 0.853\tiny{\emph{0.002}} & 0.623\tiny{\emph{0.004}} & 0.849\tiny{\emph{0.001}} & 0.657\tiny{\emph{0.009}} & 0.869\tiny{\emph{0.003}} \\
                                                                      & Multi-Task ($\mtmodel$) & 0.698\tiny{\emph{0.008}} & 0.881\tiny{\emph{0.001}} & 0.716\tiny{\emph{0.023}} & 0.890\tiny{\emph{0.002}} & 0.667\tiny{\emph{0.006}} & 0.873\tiny{\emph{0.001}} \\
                                                                      
\bottomrule
\end{tabular}%
}
\end{table}

\subsection{Integrating \intanntagr and diagnosis prediction}

Finally, we leverage this link between malignancy and \intanntagr and investigate whether jointly learning to predict IAA improves diagnostic performance (Sec.~\ref{subsec:mtmodel_method}) by comparing diagnosis-only models ($\diagmodel$) with multi-task models ($\mtmodel$). As before, we use Smooth-$L_1$ loss for $\regloss$ and focal loss~\cite{lin2017focal} for $\diagloss$. The multi-task architecture shares a common backbone and employs two heads: a regression head (as in Sec.~\ref{subsec:m1_results}) and a classification head (\texttt{Linear(256)} $\to$ \texttt{BatchNorm1D} $\to$ \texttt{ReLU} $\to$ \texttt{Dropout(0.5)} $\to$ \texttt{Linear($n_{\textrm{classes}}$)}).
To study the impact of loss weighting, we vary $\alpha$ in Eqn.~\ref{eqn:mtmodel_loss}, assigning lower ($\alpha \in \{0.1, 0.2\}$), equal ($\alpha = 0.5$), and higher ($\alpha \in \{0.8, 0.9\}$) emphasis on the diagnosis loss $\diagloss$. All models are trained under the same setup as Sec.~\ref{subsec:m1_results}, except we select the model with the highest balanced accuracy on the validation set. We report balanced accuracy and AUROC in Table~\ref{tab:ablation_results}.
Across all architectures, we find that $\alpha = 0.9$ yields the best diagnostic performance. Moreover, multi-task models ($\mtmodel$) with equal or greater emphasis on $\diagloss$ ($\alpha \geq 0.5$) consistently outperform diagnosis-only models ($\diagmodel$), confirming our hypothesis that \intanntagr prediction serves as a beneficial auxiliary task for diagnosis.

To assess generalizability, we fine-tune the $\diagmodel$ and $\mtmodel$ models (trained on \newdatasetname with $\alpha = 0.9$ for $\mtmodel$) on external datasets: \phtwo, \dermpt, \isiceight, and \isicnine. Since these datasets lack multiple annotations and thus have no IAA labels, we freeze the regression head of $\mtmodel$ before fine-tuning. Fine-tuning is conducted for 15 epochs using SGD (momentum = 0.9, weight decay = 1e-4, batch size = 32, learning rate = 1e-3 with $\times 0.1$ decay every 3 epochs). Results in Table~\ref{tab:other_datasets} show that $\mtmodel$ outperforms $\diagmodel$ across all datasets and architectures, suggesting that the performance gains from learning to predict IAA on \newdatasetname may be transferable to new datasets.

\section{Conclusion}

We studied the problem of \intanntagr (IAA) for skin lesion segmentation, and demonstrated, through agreement metrics and statistical tests, a clear relationship between IAA and malignancy. We showed that IAA can be predicted from image content alone.
Across five dermoscopic datasets, we further showed that incorporating IAA prediction as an auxiliary task in a multi-task diagnosis model improves performance over diagnosis-only models. 
To support this study, we curated \newdatasetname, the largest publicly available multi-annotator skin lesion segmentation dataset, in terms of both the number of segmentations and unique annotators. To our knowledge, this is the most extensive IAA study in skin image analysis. Future work would assess how 
% to test for the null of benign-dominates-malignant which is stronger than rejecting malignant-dominates-benign and failing to reject benign-dominates-malignant,
to test for null of non-dominance against dominance~\cite{whang2019econometric},
\rev{evaluate other boundary-based metrics such as Hausdorff distance and boundary-F1 score,} 
explore groupwise IAA measures instead of the pairwise measures: Dice and Hausdorff distance, and
examine how inter-annotator variability impacts the ABCD score.

% \begin{comment}  %% removed for anonymized MICCAI 2025 submission.
    
    % The following acknowledgement and disclaimer sections should be removed for the double-blind review process.  
    % If and when your paper is accepted, reinsert the acknowledgement and the disclaimer clause in your final camera-ready version.

\begin{credits}
\subsubsection{\ackname}
\rev{
The authors are grateful for the computational resources provided by NVIDIA Corporation and Digital Research Alliance of Canada (formerly Compute Canada). Partial funding for this project was provided by the Natural Sciences and Engineering Research Council of Canada (NSERC RGPIN/06752-2020).
}
% A bold run-in heading in small font size at the end of the paper is used for general acknowledgments, for example: This study was funded by X (grant number Y).

\subsubsection{\discintname}
\rev{
The authors have no competing interests to declare.
}
% It is now necessary to declare any competing interests or to specifically state that the authors have no competing interests. Please place the statement with a bold run-in heading in small font size beneath the (optional) acknowledgments\footnote{If EquinOCS, our proceedings submission system, is used, then the disclaimer can be provided directly in the system.}, for example: The authors have no competing interests to declare that are relevant to the content of this article. Or: Author A has received research grants from Company W. Author B has received a speaker honorarium from Company X and owns stock in Company Y. Author C is a member of committee Z.
\end{credits}

% \end{comment}

% \clearpage

\bibliographystyle{splncs04}
% \bibliography{refs}
\bibliography{refs_full}
\end{document}